\documentclass[letterpaper]{article} % DO NOT CHANGE THIS
\usepackage{aaai2026}  % DO NOT CHANGE THIS
\usepackage{times}  % DO NOT CHANGE THIS
\usepackage{helvet}  % DO NOT CHANGE THIS
\usepackage{courier}  % DO NOT CHANGE THIS
\usepackage[hyphens]{url}  % DO NOT CHANGE THIS
\usepackage{graphicx} % DO NOT CHANGE THIS
\usepackage{enumitem}
\usepackage{natbib}  % DO NOT CHANGE THIS AND DO NOT ADD ANY OPTIONS TO IT
\usepackage{caption} % DO NOT CHANGE THIS AND DO NOT ADD ANY OPTIONS TO IT
\usepackage{algorithm}
\usepackage{algorithmic}

\usepackage{amsmath}
\usepackage{booktabs}
\usepackage{multirow}
\usepackage{makecell}
\usepackage[table]{xcolor}

\usepackage{newfloat}
\usepackage{listings}
\DeclareCaptionStyle{ruled}{labelfont=normalfont,labelsep=colon,strut=off} % DO NOT CHANGE THIS
\floatstyle{ruled}
\newfloat{listing}{tb}{lst}{}
\floatname{listing}{Listing}
\title{GREAT: Generalizable Representation Enhancement via Auxiliary Transformations for Zero-Shot Environmental Prediction}
\author{
    Shiyuan Luo\textsuperscript{\rm 1},
    Chonghao Qiu\textsuperscript{\rm 1},
    Runlong Yu\textsuperscript{\rm 2},
    Yiqun Xie\textsuperscript{\rm 3},
    Xiaowei Jia\textsuperscript{\rm 1}
}
\affiliations{
    \textsuperscript{\rm 1}University of Pittsburgh,
    \textsuperscript{\rm 2}University of Alabama,
    \textsuperscript{\rm 3}University of Maryland
    \{shl298, chq29, xiaowei\}@pitt.edu, ryu5@ua.edu, xie@umd.edu
}

\begin{document}

\maketitle

\begin{abstract}
Environmental modeling faces critical challenges in predicting ecosystem dynamics across unmonitored regions due to limited and geographically imbalanced observation data. This challenge is compounded by spatial heterogeneity, causing models to learn spurious patterns that fit only local data. Unlike conventional domain generalization, environmental modeling must preserve invariant physical relationships and temporal coherence during augmentation.
In this paper, we introduce \textbf{G}eneralizable \textbf{R}epresentation \textbf{E}nhancement via \textbf{A}uxiliary \textbf{T}ransformations (\textbf{GREAT}), a framework that effectively augments available datasets to improve predictions in completely unseen regions. GREAT guides the augmentation process to ensure that the original governing processes can be recovered from the augmented data, and the inclusion of the augmented data leads to improved model generalization. Specifically, GREAT learns transformation functions at multiple layers of neural networks to augment both raw environmental features and temporal influence. They are refined through a novel bi-level training process that constrains augmented data to preserve key patterns of the original source data. 
We demonstrate GREAT's effectiveness on stream temperature prediction across six ecologically diverse watersheds in the eastern U.S., each containing multiple stream segments. Experimental results show that GREAT significantly outperforms existing methods in zero-shot scenarios.
This work provides a practical solution for environmental applications where comprehensive monitoring is infeasible.
\end{abstract}

    % \link{Code}{https://anonymous.4open.science/r/GREAT4AISI}

\section{Introduction}

Modeling environmental ecosystems plays a critical role in supporting scientific understanding, sustainable management, and resource allocation~\cite{jasechko2024rapid, o2023cultural}. 
% It provides insights into key physical variables that govern ecosystem dynamics and are 
It is essential for addressing challenges such as food security, water safety, and biodiversity conservation. For example, 
% drinking water reservoir operators in the Delaware River Basin rely on accurate predictions of water temperature and flow to determine when to take action, both for maintaining sufficient safe drinking water for 
% over 15 million people and for maintaining sufficient cool water temperatures in the river network downstream of the reservoirs \cite{williamson2015summary}. 
drinking water reservoir operators in the Delaware River Basin rely on accurate predictions of water temperature and flow to ensure sufficient safe drinking water for over 15 million people and to maintain cool river temperatures downstream of the reservoirs~\cite{williamson2015summary}.
Developing such models is inherently complex, as it requires capturing interactions among diverse physical variables and ensuring scalability across broad spatial and temporal domains.

Recent advances in machine learning (ML) have shown significant potential for improving the modeling of complex ecological systems, with successful applications across domains such as agroecosystems~\cite{liu2022kgml,cheng2025knowledge} and freshwater ecosystems~\cite{willard2025time, yu2025physics}.  
Although ML models are capable of capturing complex patterns, their generalizability largely depends on the availability of sufficient high-quality training data that adequately represents unseen scenarios. 
In real-world environmental problems, observation data can be highly limited and localized, since consistent environmental monitoring requires expensive field study, infrastructure, and ongoing maintenance, making comprehensive observation networks economically infeasible~\cite{luo2025learning, zhuang2020comprehensive}.  
This often results in regional data imbalance, where only a few regions are well observed, while other regions are sparsely observed or completely unobserved.

The development of generalizable ML models becomes even more challenging due to the spatial heterogeneity, i.e., different regions can exhibit distinct climate patterns, land use characteristics, topography, etc.~\cite{xie2021statistically, kratzert2019towards}. 
As a result, standard data-driven models that rely on local supervision may extract features that fit only local data but fail to generalize to unseen regions~\cite{karpatne2024knowledge}.  
For example, when modeling water temperature, shallow streams tend to show strong correlation with air temperature, while such patterns may not generalize to deep waters due to water's high specific heat capacity. 
% Instead, air temperature can be used with other features to account for heat exchange, which captures a more robust relationship across regions. 
Rather than relying on such intuitive correlation, air temperature combined with other physical features can help capture heat exchange with water, which offers more robust relationships across regions.

Domain generalization (DG) offers a promising paradigm by training models on source domains to generalize to unseen target domains without requiring access to target labels~\cite{wang2022generalizing, zhou2022domain, khoee2024domain,shen2023differentiable}. Data augmentation has emerged as a critical component in DG~\cite{zhou2021metaaugment, carlucci2019domain, zhou2021domain}, showing success in computer vision and natural language processing. 
Despite the previous success of these techniques,  environmental modeling presents unique challenges.  
The temporal dynamics of environmental variables are governed by complex, domain-invariant physical processes that need to be preserved during domain generalization. Simple augmentation approaches (e.g., through perturbation) may disrupt these physical relationships and temporal coherence critical for environmental prediction. The spatial variations, in particular, arise from certain physical components (e.g., groundwater influence, canopy shading) rather than superficial changes like image style or text genre. Due to these differences, models may easily learn spurious patterns that fit only local data.  Therefore, augmentation strategies must be directed to reflect these variations while preserving the general physical relationships.

Our work addresses a critical and underexplored real-world challenge: How can we effectively augment available datasets to improve generalized predictions in completely unseen regions? The key idea is to guide the augmentation process such that (1) the original governing processes can be recovered from the augmented data, and (2) the inclusion of augmented data leads to improved model generalization, as verified using reference data.
% from other sparsely observed regions. 
% It is important to note that 
Our approach differs fundamentally from existing transfer learning and meta-learning methods. Transfer learning focuses on improving performance for a specific, known target domain, whereas our goal is to develop a single generalizable model not tailored to any particular site. Similarly, while meta-learning aims to learn how to learn or refine models for individual target domains, our approach constructs a unified model requiring no further adaptation and able to generalize to a large number of completely unobserved domains, addressing a common challenge in environmental applications.

Specifically, we propose \textbf{G}eneralizable \textbf{R}epresentation \textbf{E}nhancement via \textbf{A}uxiliary \textbf{T}ransformations (\textbf{GREAT}), a novel framework designed for zero-shot generalization across geographically diverse locations. 
Our approach strategically leverages the heterogeneous data landscape: we designate a well-monitored region as the primary source domain (providing abundant training data) and treat other sparsely monitored regions as auxiliary reference domains (providing validation signals). 
GREAT's key innovation lies in learning transformation functions at multiple neural network layers that augment primary source data to help the predictive model capture generalizable patterns, with auxiliary reference domains serving as a validation signal rather than direct training data. 
These transformations, applied to both raw features and temporal dynamics, are first initialized to promote diverse augmentations and then refined through a novel bi-level optimization process: under the constraint that augmented data must preserve key patterns of the original source data, the model tuned on augmented data is optimized to perform well on reference domains. 
This design prevents overfitting to sparse auxiliary data while ensuring transformations enhance the learning of domain-invariant patterns.

GREAT's effectiveness is evaluated on stream temperature prediction, a critical indicator of freshwater ecosystem health
% that influences species survival, water quality, and energy exchange processes
~\cite{phillips2020time}, using data from multiple ecologically diverse watersheds from the eastern United States.
% GREAT can effectively enhance domain generalization and significantly outperforms existing domain generalization methods in zero-shot prediction scenarios.
In summary, our key contributions are:
\begin{itemize}[topsep=0pt, itemsep=0pt, parsep=0pt]
    \item We identify and formalize a critical challenge in environmental modeling: leveraging heterogeneous monitoring data (combining dense and sparse observations) for zero-shot prediction in unseen regions.
    \item We propose GREAT, a novel framework that learns generalizable data augmentations through multi-layer transformation modules and bi-level optimization, using sparsely monitored domains as validation signals.
    \item We demonstrate that GREAT consistently outperforms existing methods on real-world stream temperature prediction across diverse watersheds under zero-shot scenarios, underscoring its practical impact for environmental monitoring applications where extensive data collection is economically infeasible.
\end{itemize}

\begin{figure*}
    \centering
    \includegraphics[width=0.7\linewidth]{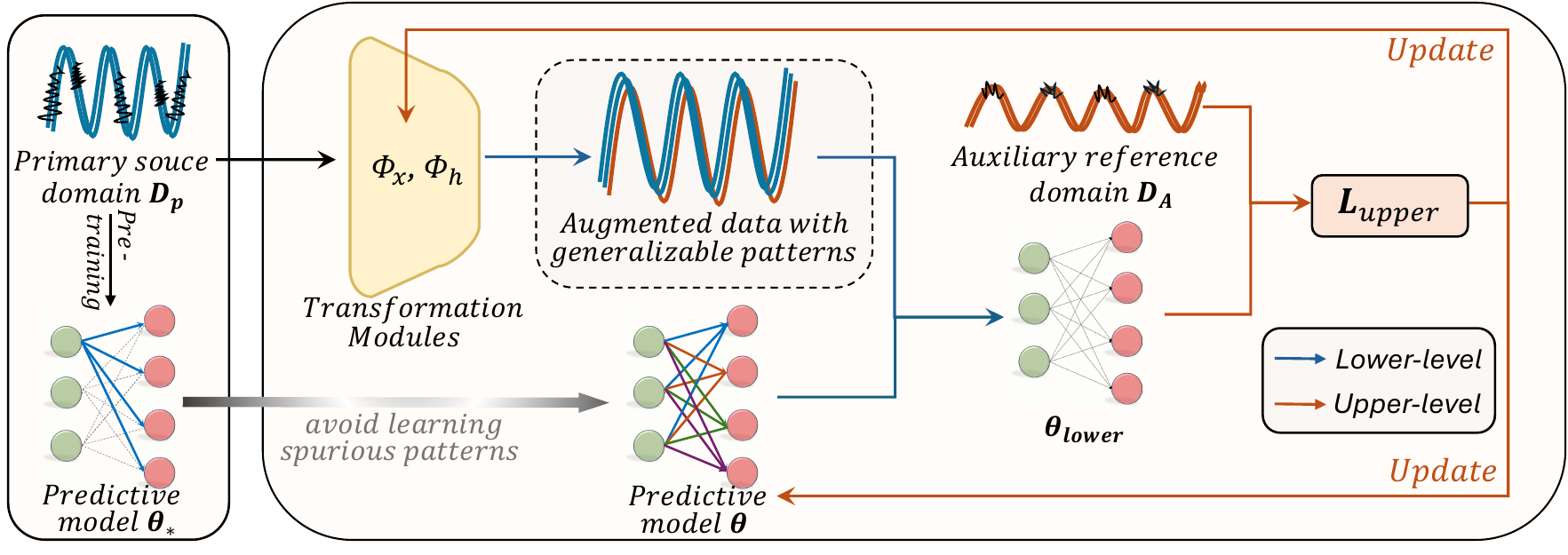}
    \caption{The overall framework of GREAT.}
    \label{fig:framework}
\end{figure*}

\section{Problem Formulation}
The goal of this work is to predict daily stream temperatures across different regions (i.e., watersheds), spanning broad geographic regions and long periods. Given a source domain with abundant temperature observations and some reference domains with limited observations, we aim to build a generalizable model that enables zero-shot prediction on completely unseen watersheds.
We formalize this as a multi-source domain generalization problem with:
\begin{itemize}[topsep=0pt, itemsep=0pt, parsep=0pt]
\item A primary source domain $\mathcal{S}_P$: One well-monitored watershed with dense temperature observations and associated input features, $\mathcal{D}_P = \{(x_i^P, y_i^P)\}_{i=1}^{N_P}$.
\item Auxiliary reference domains $\mathcal{S}_A$: Additional watersheds with limited and sparse observations, 
$\mathcal{D}_A = \{(x_i^A, y_i^A)\}_{i=1}^{N_A}$, providing complementary hydrological information.
\item Target domains $\mathcal{U}$: Completely unseen watersheds that are unavailable during training, $\mathcal{D}_U = \{(x_i^U, y_i^U)\}_{i=1}^{N_U}$.
\end{itemize}

Our tests include watersheds selected from ecologically diverse geographic regions, each watershed (in $\mathcal{S}_P$, $\mathcal{S}_A$, or $\,\mathcal{U}$) contains a set of data samples $\{(x_i, y_i)\}$, which are collected over multiple dates and from multiple locations (i.e., river segments) within the watershed. 
More formally, for each river segment $i$, we represent daily input features as $x_i = \{{x}_{i,1}, {x}_{i,2}, ..., {x}_{i,T}\}$. These features include both meteorological variables (e.g., air temperature) and physical attributes (e.g., segment width, depth). Stream temperature labels are represented as ${y}_i=\{y_{i,t}\}$, where $y_{i,t}$ denotes water temperature at location $i$ and time step $t$. Observations are sparsely available for certain locations and times, varying significantly between watersheds. 
% For example, some watersheds (e.g., in $\mathcal{S}_A$) have observations from very few locations and dates. 

\section{Method}
Real-world environmental problems often involve very few densely observed regions $\mathcal{S}_P$ (e.g., those extensively studied by research organizations) and some sparsely observed regions $\mathcal{S}_A$. Different regions can exhibit highly heterogeneous characteristics. 
Prior work in transfer learning often considers $\mathcal{S}_P$ as the source domain and  $\mathcal{S}_A$ as the target domain, e.g.,  by fine-tuning the model toward $\mathcal{S}_A$  or extracting domain-invariant features between $\mathcal{S}_P$ and $\mathcal{S}_A$.    
The typical transfer learning approaches cannot be directly used for domain generalization as they often fit only the target domain $\mathcal{S}_A$ and fail to generalize given insufficient data from $\mathcal{S}_A$. 

In this section, we present the GREAT framework, which takes a different approach to achieve zero-shot stream temperature prediction by learning generalizable patterns through feature transformations and bi-level training, as depicted in Fig.~\ref{fig:framework}. %, GREAT takes a different approach:
Rather than adapting the model directly to well-observed region $\mathcal{S}_P$ or sparsely-observed regions $\mathcal{S}_A$, we augment the dataset in $\mathcal{S}_P$ via learnable feature transformation modules. These modules are designed to neutralize domain-specific variations (marked by black symbols) so that the model, when learning from the resulting augmented data, is able to capture generalizable hydrological patterns. 
Auxiliary domains serve as a validation signal for transformation quality, ensuring the learned augmentations enhance cross-domain generalization to entirely unseen regions. %This process ensures the final predictive model learns domain-invariant relationships and can achieve strong generalization in entirely unseen regions.

The GREAT framework 
% aims to enhance model generalizability through data augmentation and 
is model-agnostic. Here we choose the long-short term memory (LSTM) model as the base model, given its widespread use in hydrology and other aquatic research~\cite{shen2021applications}. The same framework is readily applicable to other ML models. We also demonstrate in our experiment that the augmented data created by GREAT can enhance various ML models. 
In the following, we first introduce the feature transformation modules and describe how they are optimized through a bi-level training process involving both $\mathcal{S}_A$ and $\mathcal{S}_P$. We then present the pre-training procedure for both the base model and the transformation modules to ensure robust performance.

\subsection{Feature Transformation}
Environmental modeling tasks, e.g., stream temperature prediction, face a fundamental challenge when applied across large geographical regions: models trained on highly localized datasets may exhibit poor generalization to new domains due to variations in environmental conditions and geographic properties. While the underlying physical mechanisms are broadly similar across sites, their observable effects can differ substantially due to local factors.
As a result, models may overfit to site-specific patterns rather than learning the domain-invariant relationships necessary for robust prediction across regions.
To address this, our feature transformation modules are designed to extract domain-invariant patterns that generalize across watersheds, rather than characteristics tied to any single domain.

In particular, we introduce an input transformation module $g_\text{input}$ with parameters $\phi_x$:
\begin{equation}
\small
    \tilde{x}_{i,t} = g_\text{input}(x_{i,t}; \phi_x), 
\end{equation}
where $\phi_x$ evolves during training to capture domain-invariant processes such as solar radiation absorption, topographic influences that generalize across watersheds. The module is implemented as a multi-layer perceptron (MLP) trained to produce generalized representations suitable for cross-domain temperature prediction.

Environmental ecosystems, like typical dynamical systems,  also show strong temporal patterns. In the LSTM model $f$, such dependencies are captured by recursively propagating the hidden representation and latent state to the next time step. The model then integrates such temporal information with current input to make predictions, which can be represented as $y_t = f(x_t, h_{t-1})$. To further learn adaptable temporal representation, we employ a transformation module over the hidden representation of LSTM. 

The transformation of the hidden representation can be expressed as follows:
\begin{equation}
\small
    \tilde{h}_{i,t} = g_\text{hidden}(h_{i,t}; \phi_h)
\end{equation}
The transformation is applied only to the final LSTM layer to maintain the hierarchical feature extraction while adapting the highest-level temporal representations. Note that the transformation over the hidden representation is optional and introduces model dependency. We will show its effect in the ablation study.

\subsubsection{Bi-level training. }
% The goal of the transformation modules is to capture cross-domain variations while maintaining generalizable representations.  
The goal of the transformation modules is to expose the model to cross-domain variations, thereby enabling it to learn representations that generalize well to unseen domains.
To achieve this, we employ a bi-level training approach:
auxiliary reference data guide the learning of transformation, while abundant primary source data ensure that the augmentations remain grounded and prevent overfitting to auxiliary domains. 
Specifically, the bi-level training consists of: (1) \textit{\textbf{lower-level}}: transforming $\mathcal{D}_P$ and training the LSTM using the transformed and original data, and (2) \textit{\textbf{upper-level}}: evaluating and refining transformation modules based on the model performance on the reference domains $\mathcal{S}_A$ and the ability to recover source data patterns.

\textit{\textbf{Lower-level}: Transformed primary source domain adaptation. }
In this level, we update LSTM parameters while keeping the transformation modules fixed. For each data point in the primary source domain $\mathcal{S}_P$, we first apply the transformation over the forward pass of the predictive model $f$ (i.e., LSTM), as follows:
\begin{equation}
\small
{\tilde{y}_{i,t}} = f(g_\text{input}(x_{i,t}), g_\text{hidden}(h_{i,t-1}); \theta).  
\end{equation}
Given current LSTM parameters $\theta$ and fixed transformation parameters ${\phi_x, \phi_h}$, the lower-level training aims to update the model $f$ using both the predictions made on the original $\mathcal{D}_P$ and the predictions after the transformation, which can be represented as follows:  
\begin{equation}
\small
    \mathcal{L}_\text{lower}(\theta; \mathcal{D}_P, \phi_x, \phi_h) \!= \!\!\!\!\!\sum_{(i,t)\in \mathcal{D}_P}\!\!\!\!\! \frac{(y_{i,t}-\hat{y}_{i,t})^2 + \lambda (y_{i,t}-\tilde{y}_{i,t})^2}{|\mathcal{D}_P|}, 
\end{equation}
where $\hat{y}$ and $\tilde{y}$ denote the forward prediction from  $x_{i,t}$ without and with using the transformation modules, respectively. The hyperparameter $\lambda$ controls the relative weight of the loss between the original and transformed predictions.

To make the bi-level training differentiable, we approximately solve the lower-level optimization using the gradient descent approach. 
In particular, the one-step gradient descent for updating the current parameters $\theta = \theta_0$ can be expressed as follows:
\begin{equation}
\small
    \theta_\text{lower} = \theta_0 - \alpha\nabla_\theta \mathcal{L}_\text{lower}(\theta; \mathcal{D}_P, \phi_x, \phi_h)
    \label{eq:lower}
\end{equation}
where $\alpha$ denotes the step size or the learning rate for gradient descent. The gradient descent process can also be extended to multiple steps. 

By using the transformation modules, GREAT incorporates diverse data with variations in both input features and temporal dynamics. By training the model to fit both the original source data and the transformed data, this approach mitigates the risk of learning spurious patterns that only fit the original source data $\mathcal{D}_P$. 

{\textit{\textbf{Upper-level}: auxiliary reference domain guided updates. }}
The training at the upper level addresses the key question: ``Which transformations enhance cross-domain generalization?" by using auxiliary performance as a signal to guide transformation learning. 
The parameters for the transformation modules (i.e., $\phi_x$ and $\phi_h$) and the original parameters of LSTM ($\theta_0$ in Eq.~\ref{eq:lower}) will be updated based on how effectively the model $f(\,\cdot\,;\theta_{\text{lower}})$ obtained from the lower-level training can generalize to auxiliary reference domains. 
At the same time, it is important to prevent the transformation modules from simply overfitting the reference domains. This issue can be addressed by (1) enforcing that the transformed representations preserve the underlying patterns of the original data (to be discussed next), and (2) considering a diverse set of auxiliary reference domains, details of which will be provided in the experiment section.  

Specifically, we define the upper-level training objective as follows:
\begin{equation}
\small
\begin{aligned}
        \mathcal{L}_\text{upper} (\theta_0,\phi_x,\phi_h; \mathcal{D}_A, &\mathcal{D}_P)  \!=  \!\sum_{(i,t)\in \mathcal{D}_A} \frac{(f(x_{i,t};\theta_\text{lower})- y_{i,t})^2}{|\mathcal{D}_A|}
        \\
        &+\gamma (\mathcal{L}_\text{rec} (g_\text{input}, \mathcal{D}_P) + \mathcal{L}_\text{rec} (g_\text{hidden}, \mathcal{D}_P)), 
        %(f_{\theta_{inner}}(X^A), Y^A) \\ 
        %&+\lambda_{input} \mathcal{L}_{input\_rec} + \lambda_{hidden} \mathcal{L}_{hidden\_rec}
\end{aligned}
\end{equation}
where the first term on the right-hand side denotes the mean squared prediction error of the obtained model $f(\,\cdot\,;\theta_{\text{lower}})$ on the reference data, which measures how well the model generalizes to the auxiliary validation data. 
The loss $\mathcal{L}_\text{rec}$ measures the reconstruction capacity to recover the original data in $\mathcal{D}_P$ from the transformed data. We apply this reconstruction loss on both the transformed input features and the transformed hidden representations. The reconstruction loss is designed to minimize the L2 discrepancy between the original data in $\mathcal{D}_P$ and the data reconstructed from transformed data using an MLP structure. For example, $\mathcal{L}_\text{rec} (g_\text{input}, \mathcal{D}_P)$ can be computed as follows:
\begin{equation}
\small
    \mathcal{L}_\text{rec} (g_\text{input}, \mathcal{D}_P) =  \sum_{(i,t)\in \mathcal{D}_P} \frac{\|x_{i,t} - {g}_\text{rec} \circ g_\text{input}(x_{i,t})\|_2^2}{|\mathcal{D}_P|}
\end{equation}
where $g_\text{rec}$ represents the MLP used to reconstruct the original data from the transformed data. The reconstruction loss on the hidden representation $h_{i,t}$ is defined similarly.

By employing these reconstruction losses with a relative weight of $\gamma$,  the framework is constrained to keep transformations within a meaningful range — diverse enough to improve generalization but constrained enough to avoid overfitting to a specific domain.

% \begin{figure*}
%     \centering
%     \includegraphics[width=0.7\linewidth]{img/SiteMap_AAAI_Aug1.pdf}
%     \caption{Map of six distinct and ecologically varied watersheds in the eastern United States.}
%     \label{fig:map}
% \end{figure*}

\begin{figure*}
    \centering
    \includegraphics[width=0.7\linewidth]{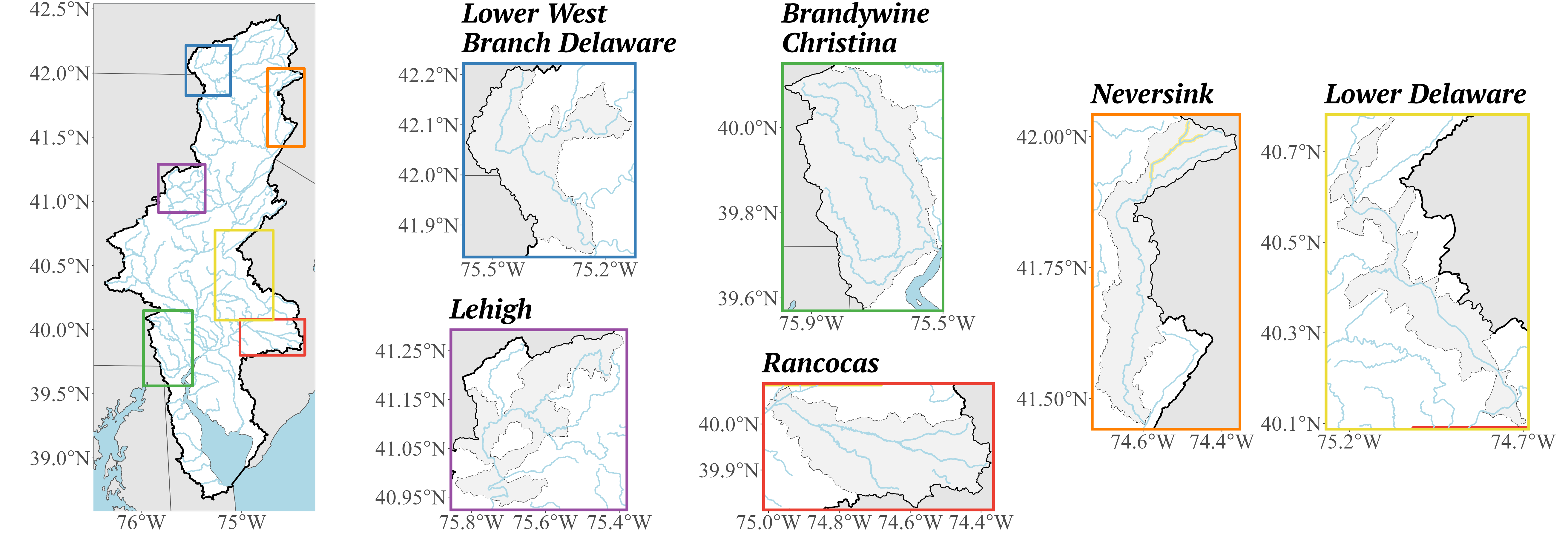}
    \caption{Map of six distinct and ecologically varied watersheds in the eastern United States.}
    \label{fig:map}
\end{figure*}

\subsubsection{Zero-shot cross-domain inference. }
After the bi-level training, the obtained predictive model can be directly applied to entirely unseen watersheds across geographically diverse regions, due to the learned transformations that capture generalizable hydrological processes. The model does not require additional local fine-tuning or calibration. 
% This zero-shot capability represents a significant advancement for practical environmental monitoring applications, where extensive data collection (e.g., sensor deployment, field surveys) over diverse and large regions is often economically infeasible.

The effectiveness of our method may be limited if the auxiliary reference data are not sufficiently representative, potentially resulting in biased validation of the predictive model. 
One possibility is to incrementally expand the dataset with additional augmented data as more auxiliary data become available, which we leave for future work. Another strategy is to pre-train the transformation modules to encourage diverse augmentations. In the following, we introduce the pre-training process for both the predictive model and the transformation modules. 

\subsection{Pre-training Process}
The pre-training stage aims to build a reasonable base predictive model to facilitate the following bi-level training and initialize feature transformation modules that are capable of generating diverse feature transformations. 

\subsubsection{Pre-training the predictive model. }
During the first iteration of lower-level training, the lack of guidance from auxiliary domain validation signals, combined with the approximate nature of the solution (Eq.~\ref{eq:lower}),  may result in the predictive model $f$ being insufficiently trained.  
Therefore, it benefits from a well-informed initialization, which ensures that the subsequent updates in the bi-level process are effective.
In general, pre-training was found to be a useful way to learn a robust model initialization~\cite{wiles2021fine}.
We establish a base predictor with initial parameters $\theta_*$ using the data-rich primary source domain by minimizing the mean squared prediction error on the source data  $\mathcal{D}_P$.

\subsubsection{Initialization of the feature transformation modules. }
Initializing transformation modules randomly or naively may result in limited diversity or trivial augmentations that do not sufficiently support the generalization of the predictive model.
To avoid this, we adopt an adversarial initialization strategy, inspired by~\cite{minderer2020automatic}, which encourages the transformation modules to generate a wide range of challenging and meaningful augmentations that serve as a crucial starting point for learning generalizable patterns in bi-level training. 
Specifically, the pre-training of transformation modules aims to minimize the loss function: 
\begin{equation}
\small
\begin{aligned}
\mathcal{L}_\text{pre} = &-\sum_{(i,t)\in \mathcal{D}_P}\frac{( f(g_\text{input}(x_{i,t}), g_\text{hidden}(h_{i,t});\theta_*) - y_{i,t})^2}{|\mathcal{D}_P|} 
\\
&+ \eta (\mathcal{L}_\text{rec}(g_\text{input},\mathcal{D}_P)+\mathcal{L}_\text{rec}(g_\text{hidden},\mathcal{D}_P). 
\end{aligned}
\end{equation}

The first term on the right-hand side is designed to increase the prediction error of the current LSTM applied exclusively to the transformed source data, thereby encouraging the discovery of patterns not yet captured by the model. The reconstruction losses, weighted by the hyper-parameter $\eta$, serve to constrain the transformation to preserve the underlying relationships needed to recover the original data.

\begin{table*}[h]
\small
\centering
\caption{Comparison of RMSE under Single and Multi settings across three sparsity levels. Lower values are better. }
\begin{tabular}{l|ccc||ccc}
\toprule
\multirow{2}{*}{\textbf{Algo.Name}} 
& \multicolumn{3}{c||}{\textbf{Single}} 
& \multicolumn{3}{c}{\textbf{Multi}} \\
\cmidrule{2-7}
& \textbf{1\%} & \textbf{0.10\%} & \textbf{0.01\%} 
& \textbf{1\%} & \textbf{0.10\%} & \textbf{0.01\%} \\
\midrule
LSTM         & \multicolumn{6}{c}{6.2475 (\textcolor{gray}{1.0007})} \\
\midrule
DroughtSet    & 6.8626 (\textcolor{gray}{0.0472}) & 7.1681 (\textcolor{gray}{0.0103}) & 7.2275 (\textcolor{gray}{0.0165}) 
             & 5.9979 (\textcolor{gray}{0.1407}) & 7.2953 (\textcolor{gray}{0.1742}) & 7.4714 (\textcolor{gray}{0.1710}) \\
MixStyle     & 5.8855 (\textcolor{gray}{0.4149}) & 6.0442 (\textcolor{gray}{0.2738}) & 6.4517 (\textcolor{gray}{0.4413}) 
             & 4.6962 (\textcolor{gray}{0.1942}) & 4.5701 (\textcolor{gray}{0.1218}) & 5.0263 (\textcolor{gray}{0.1802}) \\
Lens         & 7.5231 (\textcolor{gray}{0.1354}) & 7.4292 (\textcolor{gray}{0.0184}) & 7.4576 (\textcolor{gray}{0.1208}) 
             & 7.2732 (\textcolor{gray}{0.0074}) & 7.2437 (\textcolor{gray}{0.0027}) & 7.2277 (\textcolor{gray}{0.0039}) \\
Sharp-MAML   & 4.6218 (\textcolor{gray}{0.1128}) & 4.8691 (\textcolor{gray}{0.5387}) & 5.5780 (\textcolor{gray}{0.1586}) 
             & 5.6646 (\textcolor{gray}{0.0345}) & 5.2762 (\textcolor{gray}{0.3836}) & 5.2328 (\textcolor{gray}{0.1992}) \\
MDA          & 5.0396 (\textcolor{gray}{0.0734}) & 5.1661 (\textcolor{gray}{0.1541}) & 5.3430 (\textcolor{gray}{0.1008}) 
             & 4.5731 (\textcolor{gray}{0.0678}) & 4.9154 (\textcolor{gray}{0.0248}) & 5.0554 (\textcolor{gray}{0.1678}) \\
DANN         & 4.6434 (\textcolor{gray}{0.0280}) & 4.7169 (\textcolor{gray}{0.0186}) & 4.7327 (\textcolor{gray}{0.0205}) 
             & 4.2157 (\textcolor{gray}{0.0264}) & 4.5078 (\textcolor{gray}{0.0169}) & 4.6090 (\textcolor{gray}{0.0198}) \\
\cmidrule{1-7}
\textbf{GREAT} & \textbf{3.8306} (\textcolor{gray}{0.0166}) & \textbf{3.8104} (\textcolor{gray}{0.0155}) & \textbf{3.9026} (\textcolor{gray}{0.0145}) 
              & \textbf{3.8877} (\textcolor{gray}{0.0205}) & \textbf{4.0150} (\textcolor{gray}{0.0315}) & \textbf{4.1555} (\textcolor{gray}{0.0069}) \\
\bottomrule
\end{tabular}
\label{tab:main_results}
\end{table*}

\section{Experimental Evaluation}

\subsubsection{Data preparation.}
We use the stream water temperature data from six geographically distinct and ecologically diverse watersheds
along the eastern coast of the United States, as shown in Fig.~\ref{fig:map}. 
In particular, we represent the six watershed datasets as Lower West Branch Delaware (\textbf{LW}), Lehigh (\textbf{UL}), and Brandywine-Christina (\textbf{BC}), Rancocas (\textbf{RC}), Neversink (\textbf{NS}), and Lower Delaware (\textbf{LD}).
% Each watershed has both input drivers and target water temperature observations.   
Each watershed involves multiple connected river segments.
% (average segment length of around 10.5 km). 

The water temperature observation data are pulled from the U.S. Geological Survey's National Water Information System~\cite{usgeologicalsurvey} and the Water Quality Portal~\cite{waterquality}, the largest standardized water quality dataset for inland and coastal waterbodies. 
Daily average water temperature observations are not consistently available for every date and every segment. 
The number of temperature observations for each segment varies and could differ significantly across different watershed datasets.
For each segment and date, the input features include slope (slp), elevation (elev), width (wid), daily average air temperature (airtemp), solar radiation (rad), precipitation (precip), and potential evapotranspiration (evap). 

The complete dataset spans from October 1, 1980, to September 30, 2021, covering 41 years. 
We designate the LD dataset as our primary source domain due to its abundant observations. LD serves as the fully labeled source domain for initial pretraining across all experiments, using data from October 01, 1984, to September 30, 2010 (26 years). 
The remaining five watersheds serve as auxiliary reference domains or target domains depending on the experimental settings (in the following section). 
For auxiliary domains, we use sparse labels at three levels (1\%, 0.10\%, 0.01\%) from October 1, 1984, to September 30, 2010 (26 years), which simulates sparsely available data in real scenarios. 
At the 1\% sparsity level,  data from each watershed contains approximately 100 observations. For the 0.10\% and 0.01\% level, the number of observations decreases proportionally. 
For the target domain, we evaluate the performance on all the data from October 01, 2010, to September 30, 2021 (11 years).

\subsubsection{Baselines.}
To evaluate the effectiveness of our framework, we compare it against seven state-of-the-art baselines:
MixStyle~\cite{zhou2021domain} synthesizes new domain styles for robustness; Lens~\cite{minderer2020automatic} learns augmentations for generalization; Sharp-MAML~\cite{abbas2022sharp} uses meta-learning for cross-task generalization;  MDA~\cite{zhao2024more} aligns multiple source domains; DANN~\cite{acuna2021f} employs adversarial training for domain-invariant features; DroughtSet~\cite{tan2025droughtset} leverages geographic-spatial information; LSTM~\cite{shen2021applications} serves as the standard baseline. 
All baselines are adapted to our problem by training on the primary source domain with full labels and auxiliary reference domains with sparse labels. For most of the methods, we treat auxiliary reference domains as additional source domains. For DANN, we use some samples from auxiliary domains as the target samples. DroughtSet is provided with corresponding watershed characteristics as its geographic information inputs.

\begin{figure}
    \centering
    \includegraphics[width=\linewidth]{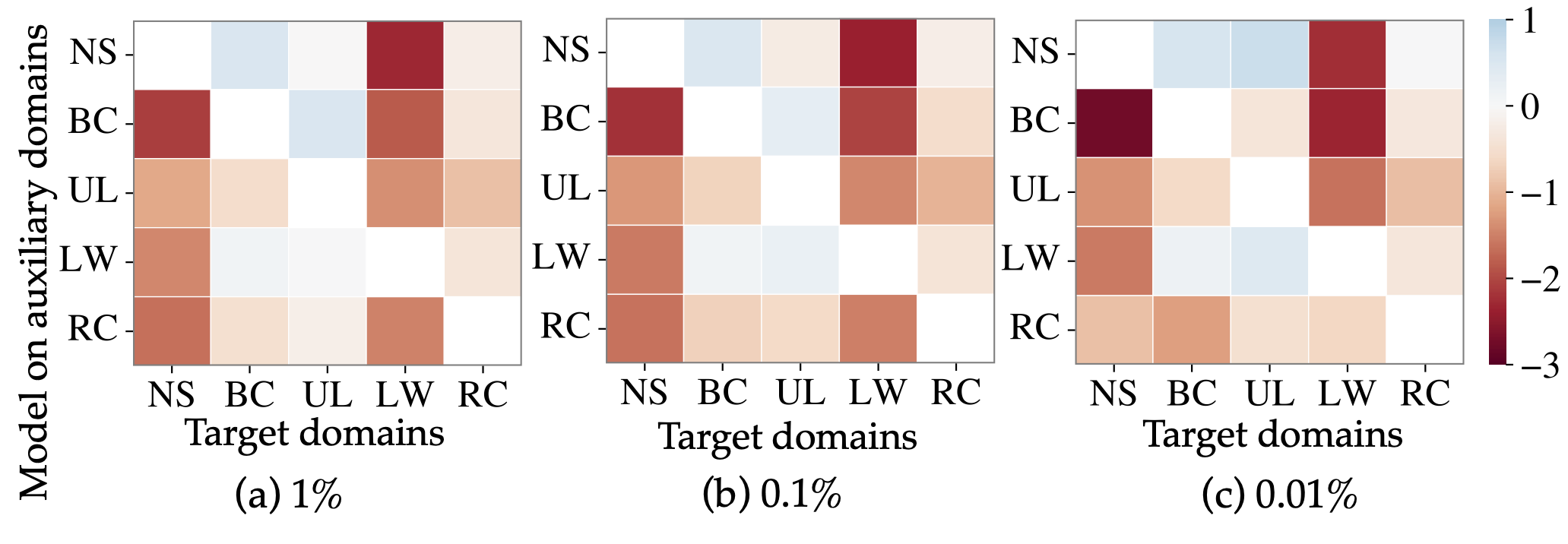}
    \caption{Difference of RMSE between GREAT and DANN. Red color means GREAT is better. }
    \label{fig:heatmap}
\end{figure}

% \subsection{Implementation Details}
% Our experiments are implemented using PyTorch with a three-layer LSTM architecture (hidden dimension 64) serving as the base temperature predictor, coupled with five-layer MLPs for feature transformation modules. The bi-level optimization framework employs an inner loop learning rate of $\alpha=0.01$ for primary watershed adaptation and an outer loop learning rate of $\gamma=0.001$ for auxiliary watershed generalization. The model utilizes balanced feature weighting ($\beta=0.5$) and reconstruction regularization with $\lambda_{\text{input}}=5.0$ and $\lambda_{\text{hidden}}=1.0$.

% Training follows a four-phase protocol: (1) LSTM pretraining for 80 epochs, (2) lens initialization for 5 epochs, (3) meta-learning for 5 epochs, and (4) final evaluation. All phases employ Adagrad optimization with a batch size of 32. The framework processes temporal sequences of 365 days with 7-dimensional input features representing climatic and physiographic characteristics. To evaluate robustness across varying data availability, we test multiple sparsity levels (0.1, 0.01, 0.001, 0.0001, $1 \times 10^{-5}$) across auxiliary watersheds. The implementation leverages the Higher library for automatic differentiation through bi-level optimization, ensuring efficient gradient tracking and real-time inference capabilities.

\subsubsection{Experiment settings.}\label{sec:setting}
We evaluate GREAT under two experimental scenarios that reflect different levels of reference data availability in environmental monitoring tasks. 
The \textbf{Single} reference domain setting simulates scenarios where limited auxiliary reference data are taken from one particular watershed, while the generalizable model is evaluated over remaining unseen watersheds. 
Here, each domain (except LD) is used in turn as the auxiliary reference domain with sparse labels, while the other four serve as target domains for zero-shot evaluation. This tests GREAT's ability to learn generalizable patterns by leveraging a single sparsely-labeled site.
% and generalize to completely unlabeled locations. 
The \textbf{Multi} reference domain setting addresses scenarios where multiple watersheds have limited data availability. Here, we use combinations of every four domains as auxiliary reference domains with sparse labels, leaving the remaining domain as target for zero-shot evaluation. This evaluates GREAT's capacity to aggregate knowledge from multiple sparsely-labeled sites for generalization.
% Detailed implementation details are provided in the supplementary file.

\begin{figure}
    \centering
    \includegraphics[width=0.9\linewidth]{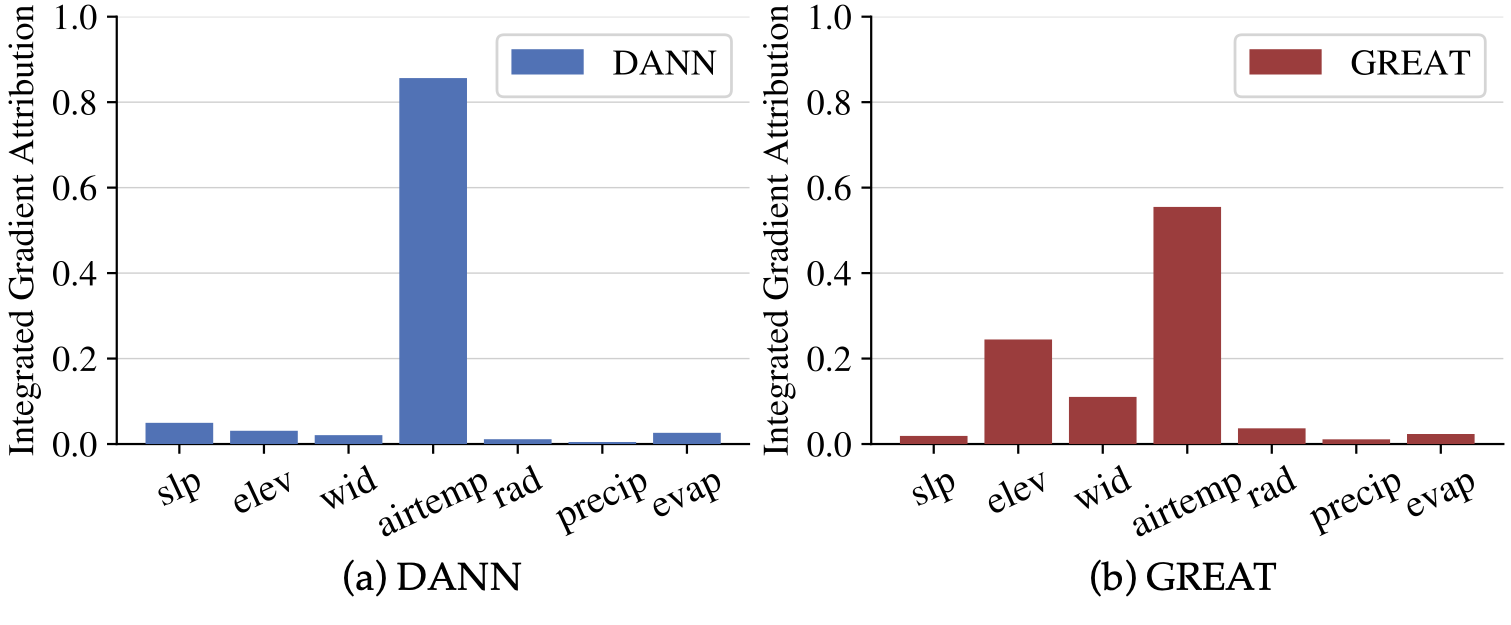}
    \caption{Comparison of feature importance attributions for DANN and GREAT.}
    \label{fig:feature}
\end{figure}

\subsubsection{Performance comparison.}
Table~\ref{tab:main_results} presents the comparative 
average performance of GREAT against baseline methods at three sparsity levels. Grey numbers denote standard deviation.
% Detailed results are provided in the supplementary file. 
We have the following observations: 
(1) The LSTM baseline, trained on one well-monitored watershed and directly applied to others, consistently performs poorly across all settings. This shows the importance of leveraging auxiliary domain information for accurate predictions. 
(2) GREAT consistently outperforms all baseline methods across different sparsity levels in both settings (Single and Multi), maintaining remarkable stability across sparsity levels, whereas baselines show performance degradation with sparser reference data. This likely results from: (i) the augmentation-based approach, which trains the predictive model primarily using the augmented data from the source domain while using the reference domain only for validation, and (ii) the incorporation of reconstruction losses, which prevent the transformation modules from overfitting to the auxiliary reference data. 
(3) Although most baselines excel in their original applications, they struggle in our environmental monitoring task because they fail to preserve physical relationships. 
Fig.~\ref{fig:feature} compares the contribution of each feature to stream water temperature prediction in entirely unseen watersheds produced by DANN and GREAT.
Feature importance was computed using the Integrated Gradients~\cite{sundararajan2017axiomatic}.
DANN's attributions tend to overfit to spurious local patterns by relying on only the correlation between air temperature and water temperature. In contrast, GREAT exhibits a more balanced and physically meaningful attribution pattern, assigning importance across both meteorological drivers and physical features, all of which contribute to the heat exchange. This demonstrates that GREAT encourages the model to focus on generalizable, physically-consistent relationships, resulting in more robust generalization to unseen regions.

To better show the performance of different target domains in the Single setting, Fig.~\ref{fig:heatmap} visualizes the performance differences between GREAT and DANN. Each cell shows the difference between GREAT and the best-performing baseline model (i.e., GREAT RMSE - DANN RMSE), with red indicating that GREAT outperforms DANN. The color intensity reflects the magnitude of improvement, and diagonal elements are left blank since those domains serve as auxiliary sources in each run. 
The consistent red coloring suggests that GREAT outperforms DANN across most domain pairs. We observe increasingly darker red cells as sparsity increases, indicating particularly strong improvements at extreme sparsity. Notably, the results also reflect that different auxiliary domains vary in their representativeness, providing different levels of validation and leading to variation in the model’s generalization performance.
% The Multi setting reveals more complex patterns due to the combinations of source domains. There are some blue cells that primarily appear for specific auxiliary reference domain combinations. However, the color of them becomes noticeably lighter at extreme sparsity, while the color of red cells become deeper, indicating GREAT's increasing advantage.

\begin{figure}
    \centering
    \includegraphics[width=0.9\linewidth]{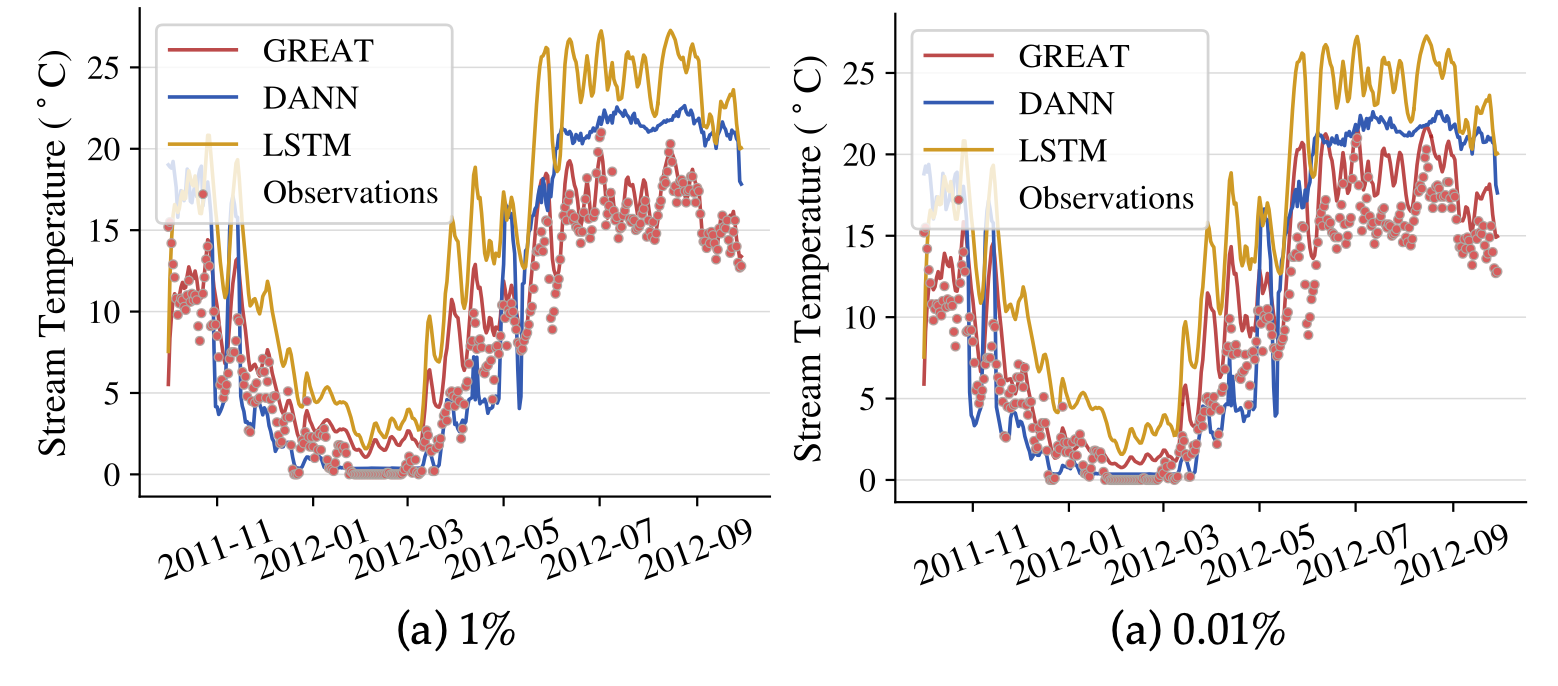}
    \caption{Time-series comparison of stream water temperature predictions by GREAT, DANN and LSTM.}
    \label{fig:timeseries}
\end{figure}

\subsubsection{Time-series analysis.}
Fig.~\ref{fig:timeseries} presents a time-series comparison of stream water temperature predictions from GREAT, DANN, and LSTM against observed data over one one year, evaluated at two different sparsity levels of auxiliary reference data: (a) 1\% and (b) 0.01\%.  The analysis reveals several insights:
% about model performance across seasonal temperature dynamics:
(1) GREAT demonstrates superior temporal modeling capabilities, aligns better with true observations compared to the other two baselines, especially in summer. Notably, GREAT maintains consistent performance even when auxiliary reference data is extremely sparse (0.01\%).
(2) LSTM consistently overestimates temperatures, particularly during summer, indicating that it fails to capture the temperature patterns from unseen domains when trained solely on the primary source domain.
(3) DANN reasonably tracks the general seasonal trend but fails to predict temperature peaks accurately, suggesting it can not preserve critical temporal patterns necessary for accurate prediction across diverse watershed conditions.
These results highlight GREAT's superior generalizability and ability to capture complex seasonal temperature patterns through its feature transformation strategy.

\begin{figure}
    \centering
    \includegraphics[width=0.9\linewidth]{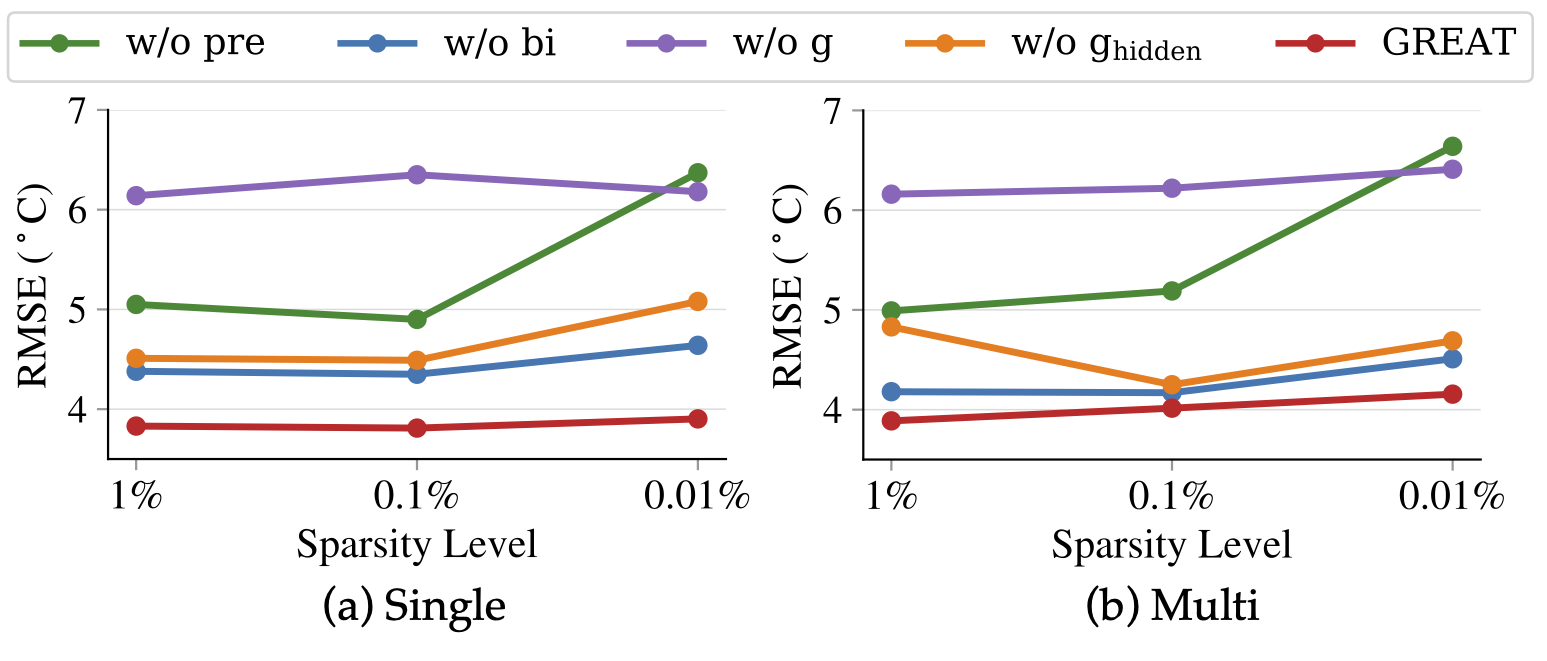}
    \caption{Results of ablation study, demonstrates the impact of different components on the overall performance GREAT.}
    \label{fig:ablation}
\end{figure}

\subsubsection{Ablation study.}
To understand the contribution of each component in GREAT, we conduct ablation studies with four variants: 
(i) w/o $\text{pre}$: removing the pre-training process.
(ii) w/o $\text{bi}$: removing the bi-level training, using the standard joint training instead while keeping both transformation modules.
(iii) w/o $g$: removing both feature transformation modules.
(iv) w/o $g_\text{hidden}$: removing the hidden state transformation ($g_\text{hidden}$), retaining only input transformation ($g_\text{input}$).

The results for comparing GREAT with these variants are shown in Fig.~\ref{fig:ablation}. 
We observe that: 
(1) Removing feature transformation modules leads to the largest performance gap across all settings, which shows their importance. 
(2) The progressive improvement from w/o $g$ $\rightarrow$ w/o $g_\text{hidden}$ $\rightarrow$ GREAT demonstrates the complementary design of multi-layer transformation. Input transformation alone provides substantial gains but can be unstable at extreme sparsity, while adding hidden state transformation stabilizes performance by capturing temporal dynamics, achieving the best performance. 
(3) Removing the pre-training process increases RMSE in both settings, especially at extreme sparsity, emphasizing the value of pre-training the predictive model and initializing the transformation modules for effective learning under limited data.
% (4) Comparing w/o $\text{bi}$ with GREAT shows that 
Bi-level training is especially crucial for Single settings, where limited auxiliary diversity makes the model prone to overfitting without careful bi-level optimization.  In contrast, 
% in Multi settings,  the natural diversity of multiple auxiliary domains create an implicit ensemble effect and also provides more samples. 
the Multi setting benefits from the natural diversity of multiple auxiliary domains, providing an implicit ensemble effect.

\begin{figure}[!t]
    \centering
    \includegraphics[width=0.6\linewidth]{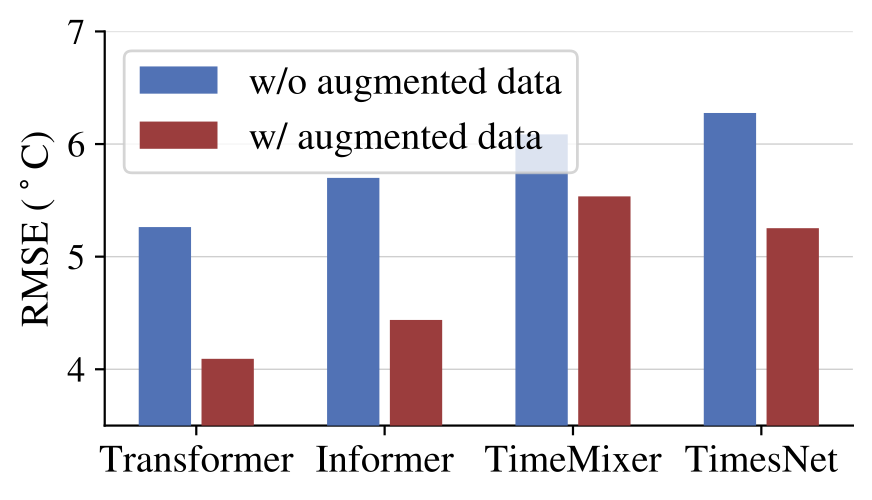}
    \caption{Impact of GREAT-augmented data on the generalization of different models.}
    \label{fig:bar}
\end{figure}

\subsubsection{Augmented data generalizability.}
This analysis aims to show whether GREAT-augmented data can enhance various ML models. 
Four state-of-the-art models are evaluated: Transformer~\cite{vaswani2017attention}, Informer~\cite{zhou2021informer}, TimeMixer~\cite{wang2024timemixer}, TimesNet~\cite{wu2022timesnet}. 
Each model is trained on the original primary source domain data (w/o augmented data) and evaluated on the remaining domain datasets, then compared to the same model trained with GREAT-augmented data (w/ augmented data).
Figure~\ref{fig:bar} shows that all models show substantial performance improvements when trained with augmented data compared to using the original data only. This indicated that our augmented data, generated by feature transformation, successfully captures generalizable patterns. These results validate that GREAT has broad applicability and the augmented data itself can serve as high-quality training data for environmental modeling, where data scarcity limits model development.

\section{Conclusion}
In this paper, we introduced a framework for zero-shot environmental prediction by enhancing a generalizable representation via auxiliary transformations. 
% GREAT learns to augment data by capturing generalizable hydrological patterns.
% Unlike previous transfer learning and domain generalization methods, 
% GREAT applies transformation modules at multiple layers of neural networks to augment primary source data and leverages sparsely observed auxiliary domains as the validation signal.
GREAT augments data by capturing generalizable hydrological patterns through transformation modules applied at multiple network layers, using sparsely observed auxiliary domains for validation.
Evaluations on stream temperature prediction across six ecologically diverse watersheds show that GREAT consistently outperforms leading baselines. Further analysis shows that
% the augmented data generated by GREAT can benefit a wide range of ML models. 
GREAT paves the way for robust environmental modeling across a range of applications in scientific domains.

\section{Acknowledgments}
This work was supported by the National Science Foundation (NSF) grants 2239175, 2316305, 2147195, 2425844, 2425845, 2430978, 2126474, 2530609, 2530610, 2203581 and 2213549; the USGS awards  G21AC10564 and G22AC00266; the NASA grants 80NSSC24K1061 and 80NSSC25K0013; and the NSF NCAR's Derecho HPC system. This research was also supported in part by the University of Pittsburgh Center for Research Computing through the resources provided. 

\bibliography{reference}

\end{document}